\documentclass[10pt,twocolumn,letterpaper]{article}

\usepackage{cvpr}
\usepackage{times}
\usepackage{epsfig}
\usepackage{graphicx}
\usepackage{amsmath}
\usepackage{amssymb}
\usepackage{float}

\usepackage[breaklinks=true,bookmarks=false]{hyperref}
\usepackage{bm}
\usepackage{multirow}
\usepackage{booktabs}
\usepackage{graphicx}

\hypersetup{
colorlinks=true,
linkcolor=red
}

\cvprfinalcopy % *** Uncomment this line for the final submission

\def\cvprPaperID{****} % *** Enter the CVPR Paper ID here

\begin{document}

%%%%%%%%% TITLE
\title{TS$^{4}$Net: Two-Stage Sample Selective Strategy for Rotating Object Detection}

\author{Kai Feng, Weixing Li, Jun Han, Feng Pan, Dongdong Zheng\\
Beijing Institute of Technology, Beijing, China\\
{\tt\small fengkai$\_$bit@outlook.com}
}

\maketitle
%\thispagestyle{empty}

%%%%%%%%% ABSTRACT
\begin{abstract}
Rotating object detection has wide applications in aerial photographs, remote sensing images, UAVs, etc. At present, most of the rotating object detection datasets focus on the field of remote sensing, and these images are usually shot in high-altitude scenes. However, image datasets captured at low-altitude areas also should be concerned, such as drone-based datasets. So we present a low-altitude drone-based dataset, named UAV-ROD, aiming to  promote the research and development in rotating object detection and UAV applications. The UAV-ROD consists of 1577 images and 30,090 instances of car category annotated by oriented bounding boxes. In particular, The UAV-ROD can be utilized for the rotating object detection, vehicle orientation recognition and object counting tasks. Compared with horizontal object detection, the regression stage of the rotation detection is a tricky problem. In this paper, we propose a rotating object detector TS$^{4}$Net, which contains anchor refinement module (ARM) and two-stage sample selective strategy (TS$^{4}$). The ARM can convert preseted horizontal anchors into high-quality rotated anchors through two-stage anchor refinement. The TS$^{4}$ module utilizes different constrained sample selective strategies to allocate positive and negative samples, which is adaptive to the regression task in different stages. Benefiting from the ARM and TS$^{4}$, the TS$^{4}$Net can achieve superior performance for rotating object detection solely with one preseted horizontal anchor. Extensive experimental results on UAV-ROD dataset and three remote sensing datasets DOTA, HRSC2016 and UCAS-AOD demonstrate that our method achieves competitive performance against most state-of-the-art methods.
\end{abstract}

%%%%%%%%% BODY TEXT
\section{Introduction}

\begin{figure}[h]
  \centering
  \includegraphics[width=\linewidth]{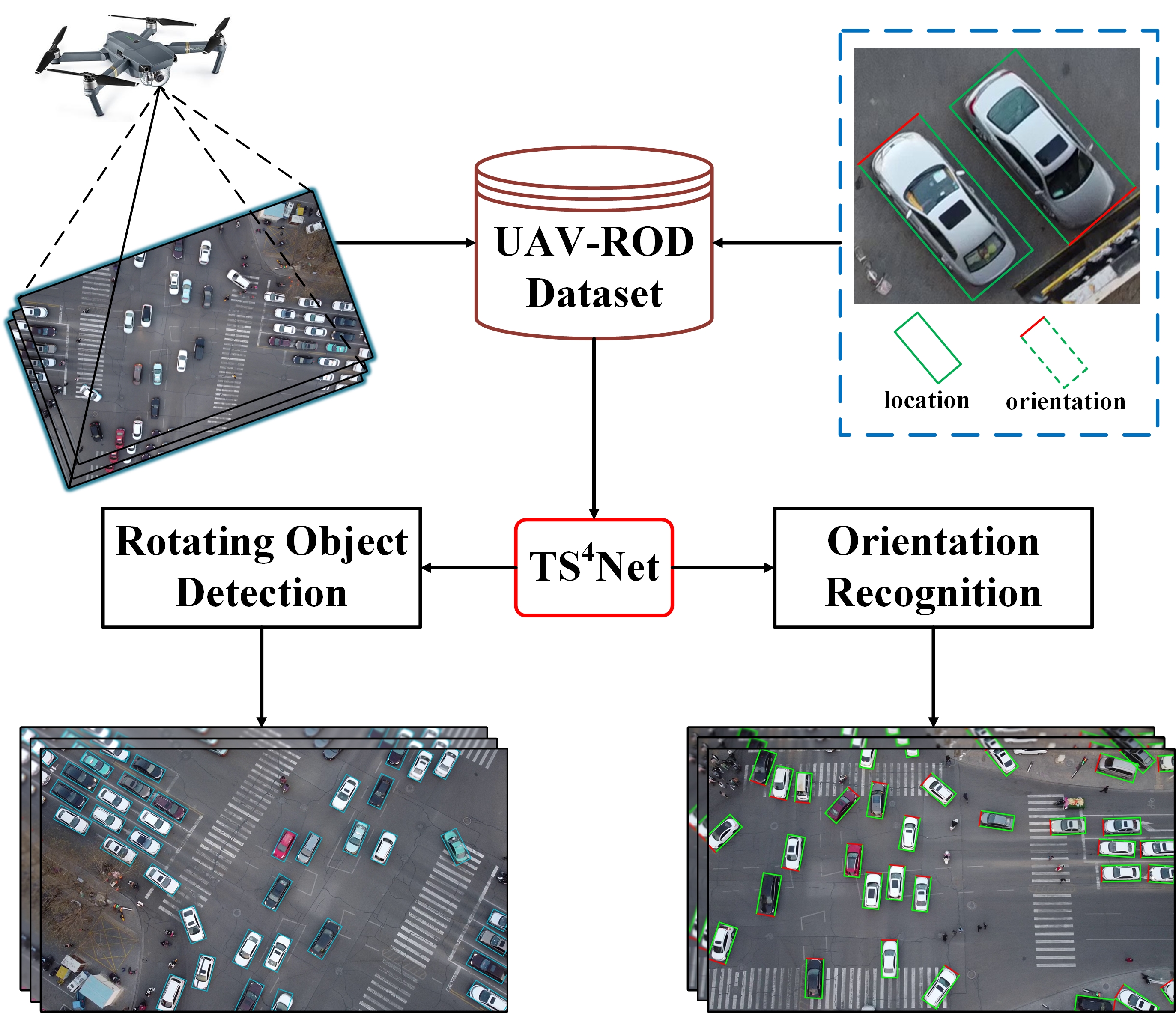}
  %\caption{Using drones to achieve rotating object detection. We propose TS$^{4}$Net to localize the cars captured by drone, and utilize the ARM and TS$^{4}$ module  to improve the detection accuracy.}
  \caption{Using drones to achieve rotating object detection and orientation recognition. We propose TS$^{4}$Net to localize the car captured by drone, and utilize the ARM and TS$^{4}$ module to improve the detection accuracy.}
  \label{fig2}
\end{figure}
Object detection is one of the most fundamental and challenging tasks in computer vision. In recent years, with the development of deep convolutional neural networks (CNN), many high-performance object detectors have been proposed \cite{DBLP:conf/nips/RenHGS15,DBLP:conf/eccv/LiuAESRFB16,DBLP:conf/cvpr/RedmonDGF16,DBLP:conf/cvpr/RedmonF17,DBLP:conf/iccv/LinGGHD17,DBLP:conf/mm/WuSWZY20}. In general, the current popular detectors can be divided into two categories: two-stage detectors and one-stage detectors. The accuracy of two-stage methods is higher, while the speed of one-stage methods is faster. These mainstream methods adopt horizontal bounding boxes to locate the position of the object detection results. However, horizontal detection suffers from many application limitations in some fields, such as drone-based object detection and remote sensing object detection. The intervention of the object angle also creates difficulties for the post-processing of the detector, such as non-maximum suppression (NMS).

Rotating object detection aims to identify the category of a specific object in image (e.g., car, ship, plane, etc.), and utilizes a rotation box to locate the object (see Figure \ref{fig2}). Many researchers have proposed different rotating object detection methods based on R-CNN frameworks \cite{DBLP:conf/cvpr/DingXLXL19, DBLP:journals/tgrs/ZhangLZ19, 9001201, DBLP:conf/iccv/YangYY0ZGSF19} or one-stage method frameworks \cite{DBLP:journals/corr/abs-1908-05612, 9377550, DBLP:journals/corr/abs-2012-04150}. Compared with horizontal detection, rotation detection seems more difficult to select positive samples from horizontal anchors, which should have large Intersection-over-Union (IoU) with the ground-truth. To fit with IoU matching, some literatures \cite{DBLP:journals/lgrs/LiuMC18, 8334248, DBLP:conf/icip/LiuHWY17} adopt rotated anchors as the original training samples. Utilizing the rotated anchor can improve the location accuracy, but it also inevitably increases the  computational complexity, especially for high-resolution aerial images. How to preset a few anchors or even one anchor to achieve high performance thus becomes a major research spot in rotating object detection. During training, the horizontal detection method adopts the constant IoU values to divide the positive and negative samples. But small angle changes in rotation detection can greatly affect the IoU, and thus the predefined IoU thresholds are not conducive to dividing the samples. So a suitable sample selective strategy is critical to reduce the variability between horizontal anchors and rotated ground-truths.

To solve the above problems, we propose the TS$^{4}$Net rotating object detection method based on RetinaNet. TS$^{4}$Net contains two modules: anchor refinement module (ARM) and two-stage sample selective strategy (TS$^{4}$). 
The ARM adopts two serial networks to refine the preseted anchors. In the first stage, the ARM adjusts the preseted horizontal anchors into high-quality rotated anchors through the fully convolutional sub-networks. In the second stage, the input rotated anchors are performed second refinement. In two stage regression task of the ARM, the anchor refinement strategies for different stages should be specifically considered. In the first stage, the gap between the preseted horizontal anchor and the ground-truth is too large, thus we only need to take a less restrictive sample selective strategy into consideration. However, the spatial location similarity between the refined anchor and ground-truth is high after the first stage of regression, so the more restrictive strategy should be adopted in the second stage. The TS$^{4}$ module makes the sample distribution more reasonable by using different degrees of restrictive sample selective methods, and can make the training more effective.

Considering existing rotating object detection datasets, most of the data are remote sensing images with high-altitude view, such as DOTA \cite{DBLP:conf/cvpr/XiaBDZBLDPZ18}, HRSC2016 \cite{7480356} and UCAS-AOD \cite{DBLP:conf/icip/ZhuCDFYJ15}. These datasets have greatly promoted the development of remote sensing object detection. However, there is a shortage of relevant datasets in the field of aerial photographs from drones. In the drone-based object detection, the drone can rotate at arbitrary-orient, so the drone-based rotating object detection also has important practical application value. Current object detection datasets for drones are usually annotated by horizontal bounding boxes, such as UAVDT \cite{DBLP:conf/eccv/DuQYYDLZHT18} and VisDrone2018 \cite{DBLP:conf/eccv/ZhuWDBLHWNCLLMW18}. Therefore, in order to enrich the data of drone-based rotating object detection, we present a dataset named UAV-ROD. The dataset contains 1577 aerial images and 30,090 car instances annotated by oriented bounding boxes. In the annotating process, we adopt the head orientation of car as the angle direction, so the orientation recognition task can also be realized in UAV-ROD. The dataset analysis is detailed in Section \ref{sec3}.

Our main contributions are summarized as follows:
\begin{itemize}
\item We propose a simple and efficient rotating object detector TS$^{4}$Net. Compared with other dense detectors, TS$^{4}$Net just needs to preset one anchor to achieve superior results.
\item For the regression problem from the horizontal anchor to the rotated ground-truth, we propose a two-stage sample selective strategy (TS$^{4}$) with two-stage anchor refinement module (ARM). Benefiting from ARM and TS$^{4}$ module, the accuracy of TS$^{4}$Net has greatly improved compared with baseline.
\item We present a rotating object detection dataset, named UAV-ROD. Compared with other drone-based datasets, UAV-ROD can be used for a variety of visual tasks, such as rotating object detection, vehicle orientation recognition and object counting\footnote{The images and annotations of UAV-ROD is available at \url{https://github.com/fengkaibit/UAV-ROD}}.
\item We achieve a competitive performance against most state-of-the-art methods on UAV-ROD dataset and three remote sensing datasets DOTA, HRSC2016 and UCAS-AOD.
\end{itemize}

\section{Related Work}
\subsection{Aerial Image Datasets}
In recent years, many high-quality aerial image datasets have been proposed for computer vision tasks. Hsieh et al. \cite{DBLP:conf/iccv/HsiehLH17} proposed a dataset named CARPK collected by drones in different parking lots, which contains 1448 images and 89,777 annotated cars. Zhu et al. \cite{DBLP:conf/eccv/ZhuWDBLHWNCLLMW18} proposed a drone-based benchmark, named VisDrone2018, consisting of 10,209 images and 263 videos with horizontal bounding box annotations. Xia et al. \cite{DBLP:conf/cvpr/XiaBDZBLDPZ18} proposed the DOTA dataset, which contains 2806 remote sensing images, and the average size of image is about 4000\time4000 pixels. The DOTA dataset contains 15 different categories and about 400,000 instances, annotated by horizontal bounding boxes and oriented bounding boxes, respectively. In \cite{7480356}, a dataset named HRSC2016 for ship detection was proposed, which includes 1061 aerial images and about 3000 annotated ships. Zhu et al. \cite{DBLP:conf/icip/ZhuCDFYJ15} proposed the UCAS-AOD dataset, consisting of 1510 remote sensing images and two categories including plane and car. Cheng et al. \cite{DBLP:journals/tgrs/ChengZH16} proposed a multi-class geospatial dataset, named NWPU VHR-10. It consists of 800 images and 10 categories, and a total of 3651 instances are annotated by horizontal bounding boxes. Long et al. \cite{DBLP:journals/tgrs/LongGXL17} proposed the RSOD dataset, which contains 976 remote sensing images and 4 categories. Du et al. \cite{DBLP:conf/eccv/DuQYYDLZHT18} proposed a large scale vehicle detection and tracking dataset captured by drones, named UAVDT, which contains about 80,000 representative video frames, 10 hours of original video and 14 attributes of vehicles, such as weather conditions, flight height, vehicle category, horizontal bounding box and occlusion.

In summary, existing drone-based datasets are deficient in rotating annotation, so the UAV-ROD dataset focuses on drone-based detection and rotation detection tasks when collecting data. We proposed the UAV-ROD dataset as a benchmark to help evaluate the performance of rotating object detection algorithms. The detailed comparison of the UAV-ROD with other datasets is presented in Table \ref{tab1}.

\subsection{Object Detection in Aerial Images}
Remote sensing and aerial photography are the main application scenarios of the rotation detection. In these scenarios, the objects are usually densely distributed, with a large aspect ratio and arbitrary angle. The current mainstream rotation detection methods can be divided into two categories like horizontal object detection: two-stage detectors and one-stage detectors. The two-stage detectors consist of two parts: a region proposal network (RPN) and R-CNN network. The RPN network uses the preseted horizontal anchors to generate high-quality region of interests (RoIs), and then the R-CNN network continues to classify and regress them. Due to the multi-stage processing and the RoI Pooling methods for feature cropping, the two-stage methods still occupy a leading position in terms of accuracy. Representative detectors are ROI-Transformer \cite{DBLP:conf/cvpr/DingXLXL19}, ICN \cite{DBLP:conf/accv/AzimiVB0R18}, SCRDet \cite{DBLP:conf/iccv/YangYY0ZGSF19}, R$^{2}$CNN \cite{DBLP:journals/corr/JiangZWYLWFL17}, etc. However, these model structures are more complex, resulting in slow inference speed. The one-stage methods achieve classification and regression through fully convolutional networks,  which is advantageous in terms of model structure and inference speed. But their accuracy still trails that of two-stage methods. Focal Loss \cite{DBLP:conf/iccv/LinGGHD17} lowers the weight of easy samples to avoid imbalance problems. So the one-stage method can also match the accuracy of the two-stage method after integrating the Focal Loss. Therefore, many one-stage rotating object detection algorithms have been proposed, such as R$^{3}$Det \cite{DBLP:journals/corr/abs-1908-05612} and S$^{2}$ANet \cite{9377550}. The above methods achieve state-of-the-art accuracy, but ignore analyzing the anchor selective strategies in different training stages. So the selected candidate samples seem to be unsatisfying, which are not favorable for model training.

\subsection{Label Assignment}
Most detectors select positive and negative samples through a specific strategy, which is called label assignment. The commonly used label assignment method is to divide the positive and negative samples by the IoU between the anchor box and the ground-truth, which is called the Max-IoU method. For instance, RetinaNet selects positive samples with IoU values higher than 0.5, and those whose IoU values lower than 0.4 are considered to be negative samples. Dynamic R-CNN \cite{DBLP:conf/eccv/ZhangCMWC20} adjusts the IoU thresholds at different stages to select high-quality positive samples dynamically. Cascade R-CNN \cite{DBLP:conf/cvpr/CaiV18} sets different IoU thresholds for different cascade detection heads to select high-quality positive samples. The Max-IoU method selects the samples by measuring the spatial relationship between the preseted anchor and the ground-truth, and has strong spatial constraints to the candidate samples. In rotating object detection, the ground-truth has an arbitrary angle, and the angle change will greatly affect the IoU. Therefore, the Max-IoU method is not conducive to selecting a large number of high-quality candidate samples in rotation detection task.

In recent years, with the emergence of the anchor-free method, the label assignment strategy based on central distance relationship has been proposed. FCOS \cite{DBLP:conf/iccv/TianSCH19} divides positive and negative samples by measuring the spatial-scale limitation and whether the anchor point falls into the ground-truth or not. ATSS \cite{DBLP:conf/cvpr/ZhangCYLL20} analyzes the main differences between FCOS and RetinaNet and concludes the sample selective strategy is the main reason for the difference of the accuracy. In this way, a label assignment strategy based on the center distance constraints is proposed. ATSS selects the candidate samples through the center distance relationship and determines the positive threshold from the statistical information. The label assignment strategy based on the distance constraints measures the importance of the sample through the center distance information, and can obtain a large number of candidate positive samples. However, in rotating object detection, this label assignment strategy does not consider the angle information, so it is a less restrictive sample selective strategy, which is not conducive to further assigning high quality samples.

\section{UAV-ROD Dataset}
\label{sec3}
\subsection{Image Collection}
Compared with remote sensing and high-altitude aerial photography datasets, we use drone to collect images in low-altitude. The flying height of the drone is between 30 meters and 80 meters. The image scenes include city roads, parking lots, residential areas, roadsides, etc. UAV-ROD dataset contains 10 aerial videos in different scenarios, and we compose the dataset by extracting images at 30 frame intervals. Finally, the UAV-ROD dataset has 1577 images whose resolution is 1920\time1080.

\begin{table*}
\centering
  \caption{Comparison of aerial image datasets. Oriented BB stands for oriented bounding box.}
  \label{tab1}
  \begin{tabular}{ccccccc}
    \toprule
    Dataset & Scenario & $\#$Images & $\#$Categories &$\#$Instances &Image width & Oriented BB\\
    \midrule
	DOTA \cite{DBLP:conf/cvpr/XiaBDZBLDPZ18}&	aerial& 	2,806	& 15 &	188.3\emph{k}&	800$\sim$4000& \bm{$\checkmark$} \\
	HRSC2016 \cite{7480356}	&aerial &	1,061	&1	&2.9\emph{k}	&$\sim$1100&	\bm{$\checkmark$}\\
	UCAS-AOD \cite{DBLP:conf/icip/ZhuCDFYJ15}	&aerial& 	1,510	&2	&14.6\emph{k}	&$\sim$1000	&\bm{$\checkmark$}\\
	NWPU VHR-10 \cite{DBLP:journals/tgrs/ChengZH16} 	&aerial&	800&	10&	3.7\emph{k}	&$\sim$1000\\	
	RSOD \cite{DBLP:journals/tgrs/LongGXL17}	&aerial&	976&	4	&6.9\emph{k}&	800$\sim$1100\\
	UAVDT \cite{DBLP:conf/eccv/DuQYYDLZHT18}	&drone&	80,000&	3&	841.5\emph{k}&	$\sim$1100\\
	VisDrone \cite{DBLP:conf/eccv/ZhuWDBLHWNCLLMW18}&	drone&	10,209&	10&	542\emph{k}&	$\sim$1920\\
	CARPK \cite{DBLP:conf/iccv/HsiehLH17} & drone &1,448 &1 &89.8\emph{k} &1280\\
	\midrule
	UAV-ROD (ours)	&drone&	1,577&	1	&30.0\emph{k}&	$\sim$1920	&\bm{$\checkmark$}\\
  \bottomrule
\end{tabular}
\end{table*}

\subsection{Data Annotation}
In the field of visual object detection, the typical method of annotation mainly adopts horizontal bounding boxes to locate the object position on image. The horizontal bounding box is marked with ($x_{c}$, $y_{c}$, $w$, $h$), where ($x_{c}$, $y_{c}$) is the center of the object, and $w$, $h$ are the width and height of the box. Some methods are also annotated with ($x_{min}$, $y_{min}$, $x_{max}$, $y_{max}$) format, where ($x_{min}$, $y_{min}$) and ($x_{max}$, $y_{max}$) are the top-left and the bottom-right corner point of the object, respectively. Horizontal annotations are widely used in natural scene images. However, in aerial object detection scenes, using horizontal annotation may cause inaccuracies in object position. Especially when the object has an arbitrary angle, the horizontal box does not fit the contour of the object closely. So we adpot the oriented bounding box for annotation.

\begin{figure}[h]
  \centering
  \includegraphics[width=\linewidth]{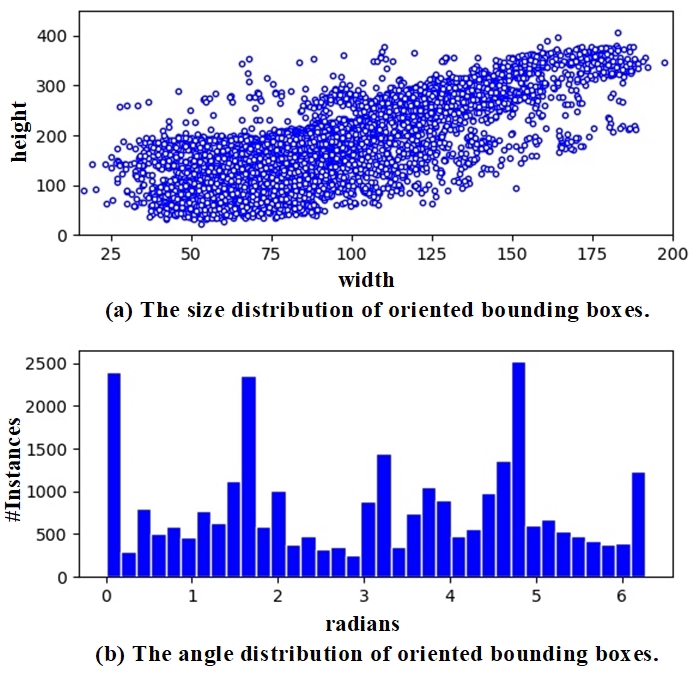}
  \caption{Schematic diagram of the size and angle statistics of UAV-ROD dataset.}
  \label{fig5}
\end{figure}

\begin{figure*}
  \centering
  \includegraphics[width=\linewidth]{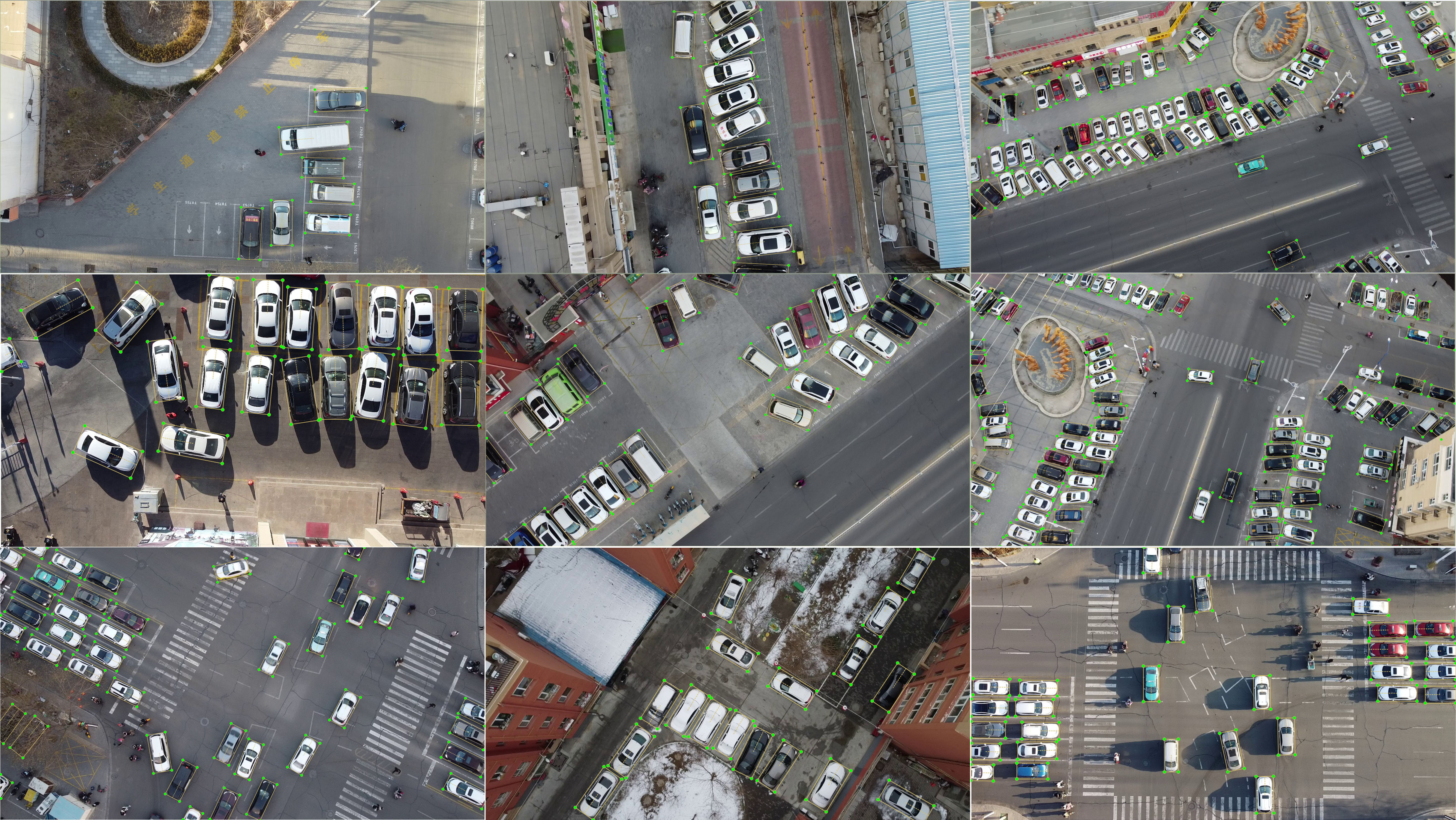}
  \caption{Some example annotated images of the UAV-ROD dataset.}
  \label{fig3}
\end{figure*}

There are two main annotation methods used for rotating object annotation: four-point annotation method and oriented bounding box method. The four-point annotation method is used in datasets such as DOTA \cite{DBLP:conf/cvpr/XiaBDZBLDPZ18} and UCAS-AOD \cite{DBLP:conf/icip/ZhuCDFYJ15}. In detail, the annotation form of this method is \{($x_{i}$, $y_{i}$), $i$=1,2,3,4\}, where ($x_{i}$, $y_{i}$) denotes the vertex position of the object. The $\theta$-based oriented bounding box is adopted in HRSC2016 \cite{7480356} and some text detection benchmark \cite{DBLP:conf/cvpr/YaoBLMT12}, and its representation form is ($x_{c}$, $y_{c}$, $w$, $h$, $\theta$). In UAV-ROD,  $\theta$ denotes the angle of car head direction and is increased in a clockwise manner. The $w$, $h$ represent the width and height of the oriented box respectively. Figure \ref{fig5} shows the statistics of the annotated size and angle of the instances.

\subsection{Data Statistics}
The UAV-ROD dataset consists of 1577 images captured by drone, and 30,090 car instances annotated with the oriented bounding box method. The average number of objects per image in our dataset is 19.08, of which the maximum number of objects per image is 104. In the image with dense objects, we also annotate all the instances in the dense area one by one. In contrast, dense objects are usually not annotated one-by-one in natural scene datasets due to the occlusion. The UAV-ROD dataset can be used in rotating object detection, vehicle orientation recognition and object counting tasks. Some examples of annotated images in our dataset are shown in Figure \ref{fig3}.

\section{Method}

\subsection{RetinaNet as Baseline}
\begin{figure}[h]
  \centering
  \includegraphics[width=2in]{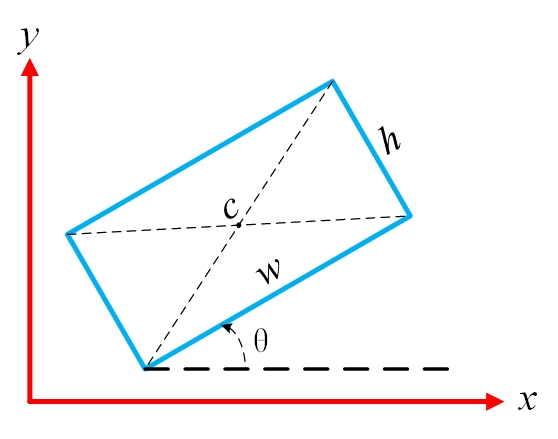}
  \caption{Oriented bounding box definition in TS$^{4}$Net. $c$ denotes the center point. $\theta$, $w$ and $h$ represent the angle, long side and the short side of bounding boxes respectively.}
  \label{fig6}
\end{figure}

We adopt the vanilla one-stage detector, RetinaNet \cite{DBLP:conf/iccv/LinGGHD17} as our baseline. It uses ResNet \cite{DBLP:conf/cvpr/HeZRS16} as the backbone, and constructs mutil-level features through feature pyramid networks (FPN) \cite{DBLP:conf/cvpr/LinDGHHB17}. Two parallel fully convolutional networks are connected after FPN to carry out the classification and regression tasks respectively. In addition, the Focal Loss is applied to overcome the imbalance problem during training. The official RetinaNet is designed for horizontal object detection, we incorporate the $\theta$ parameter in the oriented box regression and have the oriented box prediction as ($x_{c}$, $y_{c}$, $w$, $h$, $\theta$). We follow the definition in \cite{DBLP:conf/cvpr/DingXLXL19} and adopt $\theta \in [-\pi/4, 3\pi/4)$. The $w,h$ are the long and short side of the oriented box respectively. An example is shown in Figure \ref{fig6}, where $\theta$ means the angle from the x-axis to the direction of $w$. For regression task, we have:
\begin{equation}
\begin{split}
\label{eq1}
  t_x=(x-x_a)/w_a,\quad t_y=(y-y_a)/h_a\\
  t_w=log(w/w_a),\quad t_h=log(h/h_a)\quad\\
  t_\theta=\theta-\theta_a+k\pi \qquad \qquad
\end{split}
\end{equation}

\begin{equation}
\begin{split}
\label{eq2}
  t^{'}_x=(x^{'}-x_a)/w_a,\quad t^{'}_y=(y^{'}-y_a)/h_a\\
  t^{'}_w=log(w^{'}/w_a),\quad t^{'}_h=log(h^{'}/h_a)\quad\\
  t^{'}_\theta=\theta^{'}-\theta_a+k\pi \qquad \qquad
\end{split}
\end{equation}
where $(x,y),w,h,\theta$ denote the bounding box center point, width, height and angle, respectively. $x$, $x_a$ and $x^{'}$ are for the ground-truth box, anchor and predicted box, respectively (likewise for $y, w, h, \theta$). $k$ is an integer to ensure ($\theta-\theta_a) \in [-\pi/4, 3\pi/4)$. Given the ground-truth box offsets $\bm{t}=(t_x,t_y,t_w,t_h,t_\theta)$ and the predicted offsets $\bm{t^{'}}=(t^{'}_x,t^{'}_y,t^{'}_w,t^{'}_h,t^{'}_\theta)$, the regression loss is calculated between $\bm{t}$ and $\bm{t^{'}}$. For testing, the predicted offsets $\bm{t^{'}}$ is inversely solved according to Equation \ref{eq2} to obtain the predicted box.

\subsection{Anchor Refinement Module}
\begin{figure*}
  \centering
  \includegraphics[width=\linewidth]{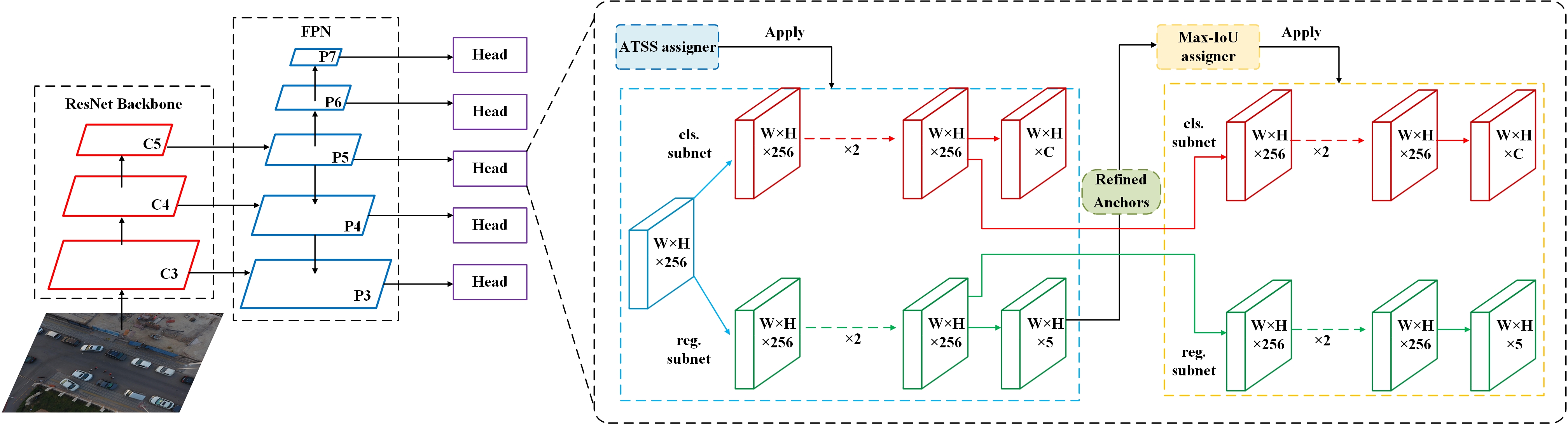}
  \caption{Architecture of the proposed TS$^{4}$Net. TS$^{4}$Net consists of a ResNet backbone, a FPN, an ARM cascade network and the two-stage sample selective strategy. `C' indicates number of categories.}
  \label{fig7}
\end{figure*}

Compared with horizontal detection, rotation detection is more difficult to regress the preseted horizontal anchors to rotated ground-truths during training. In particular, it is a challenge to select the positive samples from the horizontal anchors with large IoU values. To address this issue, we adopt a two-stage anchor refinement module (ARM) in TS$^{4}$Net (see Figure \ref{fig7}). The ARM includes a two-stage cascade network, and each network contains parallel classification and regression sub-networks, which are used for category prediction and rotated box regression respectively. The ARM first classifies and regresses the preseted anchor in the first stage and transforms the horizontal anchor into high-quality rotated anchor, facilitating the second stage of the anchor regression process. In the second stage, the rotated anchors are fed into the second subnetwork to yield more accurate prediction results. Profiting from the ARM, the single-stage detector can be optimized several times for the preseted anchor like the multi-stage detector, thus making the detection results more precise. In order to improve the inference speed, the two stages of the ARM use two 256-channel convolutional layers as classification and regression networks. During training, the total loss contains the two-stage loss of the ARM, as shown in Equation \ref{eq3}:
\begin{equation}
\small
\begin{split}
\label{eq3}
L=\frac{1}{N^{1}_{cls}}\sum_{i}L^{1}_{cls}(p_i,p^{*}_i)+\lambda\frac{1}{N^{1}_{reg}}\sum_{i}p^{*}_{i}L^{1}_{reg}(t_i,t^{*}_i)\\
	+\frac{1}{N^{2}_{cls}}\sum_{i}L^{2}_{cls}(p_i,p^{*}_i)+\lambda\frac{1}{N^{2}_{reg}}\sum_{i}p^{*}_{i}L^{2}_{reg}(t_i,t^{*}_i)
\end{split}
\end{equation}
where $L_{cls}$ and $L_{reg}$ are implemented by focal loss and smooth L1 loss, $N_{cls}$ and $N_{reg}$ are the total number of samples involved in classification and regression, respectively. The superscripts 1 and 2 respectively represent the first and second stage of the ARM. Variable $p^{*}$ represents the class label for anchors ($p^{*}=1$ for positive samples and $p^{*}=0$ for negative samples). The hyper-parameter $\lambda$ controls the trade-off and is set to 1 by default.

\subsection{Two-stage Sample Selective Strategy}
In this section, we discuss the differences between two sample selective strategies in rotation detection from different perspectives, and propose a two-stage sample selective strategy to combine their advantages.

\textbf{Sample selective strategy based on spatial information.} The Max-IoU is a typical method that uses spatial information to select samples. Specifically, an anchor is classified as a positive sample when the IoU value between the anchor and the ground-truth is greater than the positive threshold, and as a negative sample when the IoU is lower than the negative threshold. This sample selective strategy is widely used in object detection. The classical object detector enhances the IoU value by setting a lot of anchors with different sizes and aspect ratios, thus boosting the number of positive samples. However, in the rotation detection, the angle leads to a low IoU value between the horizontal anchor and the ground-truth. Moreover, the IoU value is greatly affected by the rotation angle, which causes a poor quality of the obtained positive samples. To alleviate this problem, many rotation detectors add preseted angles to the anchor, thus improving the IoU value with ground-truth. However, it will significantly increase the quantity of anchors, which will greatly aggravate the computational efficiency of the model for high-resolution aerial images.

\textbf{Sample selective strategy based on central distance information.} In FCOS and ATSS algorithms, the significance of samples is measured by the distance between the center point of the samples and the ground-truth. Compared with the Max-IoU method, the method based on the center distance is less constrained, and enables a larger number of positive samples to be obtained. Moreover, this method sets only one sample at each anchor point, which significantly reduces the computations and improves the efficiency. In the rotating object detection, the ground-truth contains the angle information of the object. However this strategy does not utilize the angle information in it, which is not conducive to further selecting the high quality candidate positive samples. In contrast, the calculation of IoU includes the relationship of angle information between the sample and the ground-truth, which is a strongly constrained strategy. Therefore, combining the advantages of the two strategies is our main motivation for proposing the TS$^{4}$ module.

\textbf{Two-stage sample selective strategy.} From the previous analysis, TS$^{4}$Net incorporates ARM to obtain more accurate localization results. The first stage of ARM refines the horizontal anchors to high quality rotated anchors, and then the second stage adjusts the rotating anchor to a more accurate prediction box. Both stages require to use the sample selective strategy. In the first stage, the detector needs to adjust the horizontal anchors to the rotated anchors. Due to the large gap between the horizontal anchor and its corresponding ground-truth, thus the less constrained ATSS sample selective strategy is adopted in first stage. The ATSS strategy divides positive and negative samples according to the statistical information of the central distance between samples and ground-truths, and obtains high quality horizontal anchors as positive samples with a moderately easy sample constraint. After the first stage of classification and regression, the horizontal anchors are refined to high quality rotated anchors. At this time, we achieve a high value of IoU and spatial similarity between the rotated anchor and ground-truth. Therefore, the model adopts Max-IoU method with strict constraint as the sample selective strategy in second stage, and obtains more accurate positive samples. The network architecture of TS$^{4}$Net is shown in Figure \ref{fig7}. The first stage of TS$^{4}$Net utilizes the horizontal anchors and ATSS strategy to complete the sample selection, and the second stage uses the rotated anchors and Max-IoU method instead. Both stages contain two parallel sub-networks for classification and regression, similar to RetinaNet.
\section{Experiments}

\subsection{Datasets}
We conduct experiments on the UAV-ROD dataset and the remote sensing datasets DOTA, HRSC2016, UCAS-AOD. The ground-truth of these datasets are all oriented bounding boxes. UAV-ROD is a drone-based car detection dataset, which contains 1150 images in the training set and 427 images in the test set. DOTA is a large remote sensing dataset, including 2806 images and 188,282 annotated instances of 15 categories. Half of the images are randomly assigned as the training set, 1/6 as the validation set and 1/3 as the test set. Since the image resolution of DOTA is too large, we crop the original images into 1024\time1024 patches with the stride of 200. HRSC2016 is a high-resolution ship recognition dataset, which contains 1061 images. The training, validation and test set on HRSC2016 include 436 images, 181 images and 444 images, respectively. UCAS-AOD is a plane and car detection dataset containing a total of 1510 images. Following the settings in the literature \cite{DBLP:conf/cvpr/XiaBDZBLDPZ18, DBLP:conf/accv/AzimiVB0R18}, we randomly select 1110 images for training and 400 images for testing. During training, the input image size of DOTA is 1024\time1024, and other dataset images are rescaled to 1333\time800.

\subsection{Implementation Details}
For the experiments, we adopt RetinaNet as baseline according to the above description. ResNet-50 is adopted as backbone and FPN is used to build the 5-level feature layers (i.e., $P_{3}$ - $P_{7}$). TS$^{4}$Net sets one square anchor box per location, with sizes \{32, 64, 128, 256, 512\} in $P_{3}$ - $P_{7}$. The total epochs of UAV-ROD, DOTA, HRSC2016 and UCAS-AOD are 12, 12, 72 and 24, respectively. We use random horizontal flipping and rotation for data augmentation. The optimizer used for training is SGD with an initial learning rate of 2.5e-3, which is divided by 10 at each decay step. The momentum and weight decay are 0.9 and 0.0001, respectively. We train the model on two V100 GPUs with a total batch size of 4.

\subsection{Ablation Study}
\begin{table}
\centering
\small
  \caption{Ablation study for TS$^{4}$Net on HRSC2016 dataset. We explore the effect of ARM, the number of anchors and the different forms of the two-stage sample selective strategy on accuracy.}
  \label{tab2}
\setlength{\tabcolsep}{1.6mm}
  \begin{tabular}{c|c|c|c|c}
    \toprule
    Method	&Assigner&	$\#$Anchor&	ARM	&mAP\\
    \midrule
	\multirow{3}{*}{RetinaNet}
    	~&Max-IoU	&27	&&	83.32\\
	~ &Max-IoU	&1	&&	82.03\\
	~&Max-IoU + Max-IoU	&1&	\bm{$\checkmark$}	&87.21\\
	\midrule
	\multirow{4}{*}{TS$^{4}$Net}
		~&Max-IoU + Max-IoU&	1&	\bm{$\checkmark$}	&87.37\\
	~ &ATSS + ATSS&	1&	\bm{$\checkmark$}	&88.12\\
	~&Max-IoU + ATSS	&1&	\bm{$\checkmark$}	&86.18\\
	~&ATSS + Max-IoU	&1&	\bm{$\checkmark$}	&\textbf{89.44}\\
  \bottomrule
\end{tabular}
\end{table}
In this section, we construct a series of experiments on HRSC2016 to verify the effectiveness of our method. All experiments adopt ResNet-50 FPN as the backbone and use RetinaNet as the baseline.

\textbf{RetinaNet Results.} To achieve rotating object detection, we integrate the angle prediction features described in subsection 4.1 into RetinaNet. And we add three angles $[0, \pi/6, \pi/3]$ at each anchor to enhance the matching effect of Max-IoU method. As shown in Table \ref{tab2}, RetinaNet achieves a mAP of 83.32{\%} on HRSC2016 and is used for our baseline. We change the number of anchor to 1, and obtain 82.03$\%$ of mAP. The result shows that few anchors also has the potential to achieve high accuracy while can reduce the computation cost of the model.

\textbf{Anchor Refinement Module.} To demonstrate the effectiveness of the ARM, we replace the original detection head with the ARM head (see the fourth row of Table \ref{tab2}). In this case, the sample selective strategy for both stages adopts Max-IoU method and the positive and negative sample thresholds are set to 0.5 and 0.4, and we obtain 87.21\% of mAP. Compared to the baseline and the previous experiment (Table \ref{tab2}, row 3), the precision is improved by 3.89\% (87.21\% $v.s.$ 83.32\%) and 5.18\% (87.21\% $v.s.$ 82.03\%) respectively. The result demonstrates that with the two-stage anchor refinement, the model can achieve high performance with only one preseted anchor.

\textbf{Two-stage sample selective strategy.} To demonstrate the correctness of the analysis in section 4.3, we design four experiments to validate the effectiveness of two-stage sample selective strategy. Firstly, we adopt Max-IoU in both stages, and the positive and negative sample thresholds in the second stage are increased to 0.6 and 0.5. The aim is to improve the quality of the sample. We obtain a mAP of 87.37\%. Changing the sample selective strategy to ATSS in both stages, the mAP is 88.12\%, which is 0.75\% higher compared to Max-IoU. Considering that the first stage is to refine horizontal anchor to rotated anchor and the second stage is the fine-tune regression of rotated anchor, we adopt the ATSS and Max-IoU methods in two sample selection processes respectively. The mAP is 89.44\%, which is higher than the previous strategies. The experimental results verify our analysis.
As a contrast experiment, we test the combination of Max-IoU as the first stage and ATSS as the second stage. We obtain a mAP of 86.18\%, which is 3.26\% lower compared to the ATSS+Max-IoU strategy. This indicates that using a strongly constrained in the first stage and a weakly constrained strategy in the second stage contradicts the process of rotating box regression, thus does not yield a credible result. The visualization of different strategies can be found in the appendix.

\subsection{Experiment Results}
In this section, we compare our proposed TS$^{4}$Net with other state-of-the-art methods on UAV-ROD, DOTA, HRSC2016 and UCAS-AOD datasets.

\textbf{Results on UAV-ROD.} On UAV-ROD dataset, our method is compared with two classic one-stage and two-stage detectors. Following the MS COCO evaluation metrics, we report the AP, AP$_{75}$ and AP$_{50}$ results. As shown in Table \ref{tab3}, we achieve the AP of 75.57\% with ResNet-50, which outperforms the baseline model by 4.11\% (75.57\% $v.s.$ 71.46\%). The results demonstrated that with ARM and TS$^{4}$, our method is able to achieve better accuracy in localization. Compared with a typical two-stage method R-Faster R-CNN, our method also achieves a comparable accuracy (75.57\% $v.s.$ 75.79\%). Using ResNet-101 as backbone, TS$^{4}$Net obtains 76.42\% AP and 98.34\% AP$_{50}$, achieving the best results.

\begin{figure}[h]
  \centering
  \includegraphics[width=\linewidth]{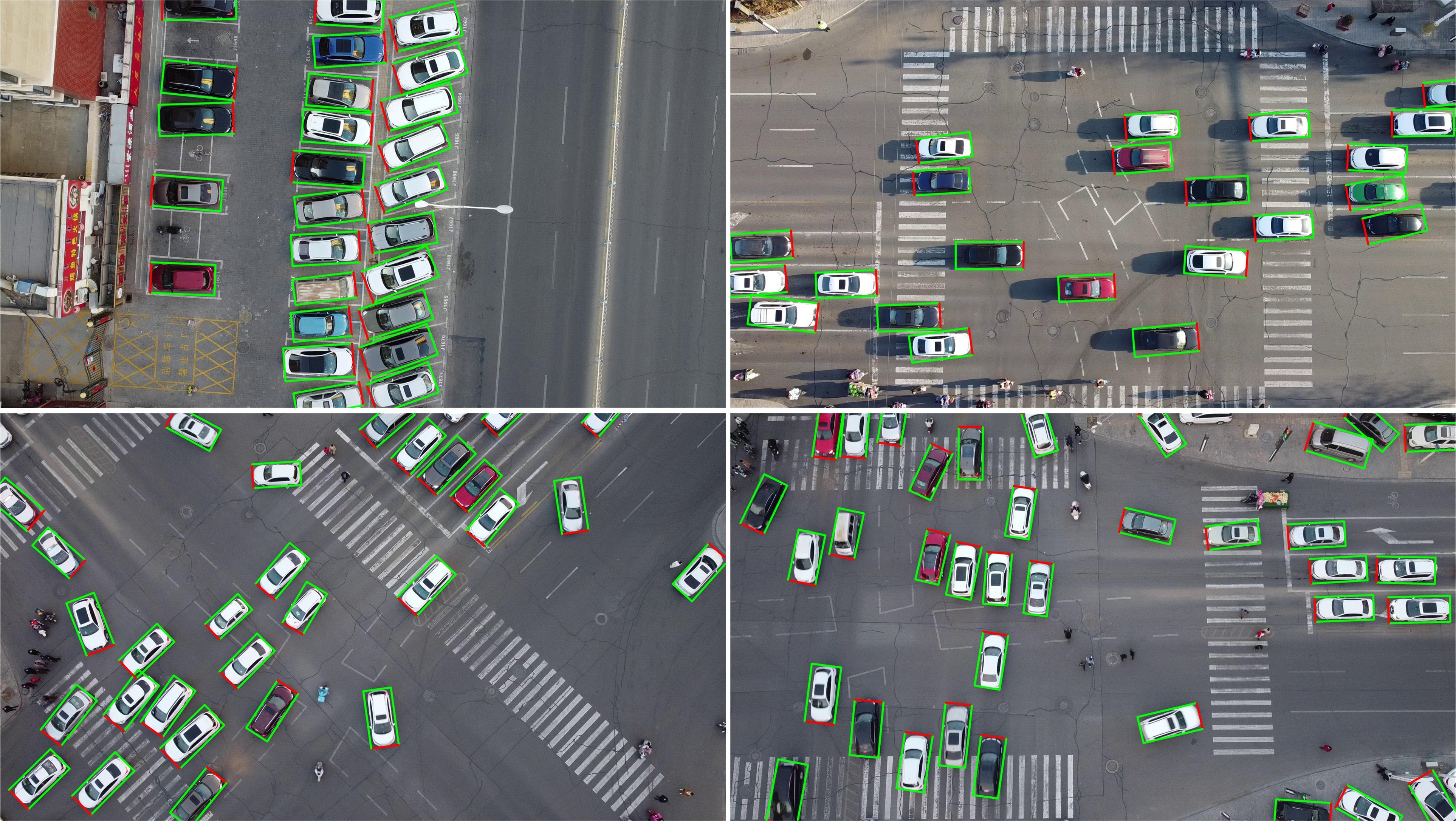}
  \caption{Visualization of car orientation recognition. The red line indicates the head direction of the car.}
  \label{head}
\end{figure}

\textbf{Results of orientation recognition task on UAV-ROD.} We adopt the UAV-ROD dataset to perform the car orientation recognition task. We modify two settings on TS$^{4}$Net: (1) The range of angle prediction is adjusted to $[0, 2\pi)$, thus can cover all head directions of the car. (2) To avoid the large offest of angle regression, the decode of angle is divided by 2$\pi$ in Equation \ref{eq1} and Equation \ref{eq2}. The predicted results of car orientation are shown in Figure \ref{head}, where the red line indicates the head direction.

\begin{table}
\centering
\small
\setlength{\tabcolsep}{2mm}
  \caption{Detection results on  UAV-ROD dataset. We report the performance in MS COCO style.}
  \label{tab3}
  \begin{tabular}{c|c|c|c|c}
    \toprule
    Method	&Backbone  &AP &AP$_{75}$ & AP$_{50}$\\
    \midrule
    R-RetinaNet \cite{DBLP:conf/iccv/LinGGHD17} &ResNet-50	& 71.46& 85.88&	97.68\\
	R-Faster R-CNN \cite{DBLP:conf/nips/RenHGS15} &ResNet-50 &75.79&86.38&98.07\\
	\midrule
	\emph{ours:}&&&&\\
	TS$^{4}$Net & ResNet-50& 75.57& 86.70& 98.03\\
	TS$^{4}$Net & ResNet-101& \textbf{76.42}& \textbf{88.39}& \textbf{98.34}\\
  \bottomrule
\end{tabular}
\end{table}

\textbf{Results on DOTA.} We compare our method with the state-of-the-arts on DOTA as shown in Table \ref{tab4}. Our method achieves mAP of 74.70\% with ResNet-50. Using ResNet-101 as the backbone, our method has reached the mAP of 75.63\% and outperforms other advanced rotation detectors. More visualization of detection results on DOTA can be found in the appendix.

\begin{table}
\centering
%\small
\setlength{\tabcolsep}{0.5mm}
  \caption{Detection results on  HRSC2016 dataset. mAP* stands for evaluated by PASCAL VOC2012 metrics.}
  \label{tab5}
  \begin{tabular}{c|c|c|c}
    \toprule
    Method	&Backbone	&\#Anchor	&mAP\\
    \midrule
    R$^{2}$CNN \cite{DBLP:journals/corr/JiangZWYLWFL17}&ResNet-101	&21&	73.07\\
	RC1\&RC2 \cite{DBLP:conf/icpram/LiuYWY17} &	VGG16	&-&	75.70\\
	RRPN \cite{rrpn}&	ResNet-101&	54	&79.08\\
	R$^{2}$PN \cite{DBLP:journals/lgrs/ZhangGZY18}&	VGG16	&24&	79.60\\
	R-RetinaNet \cite{DBLP:conf/iccv/LinGGHD17}&	ResNet-50&	27	&83.32\\
	RRD \cite{DBLP:conf/cvpr/LiaoZSXB18}&	VGG16&	13&	84.3\\
	RoI-Trans. \cite{DBLP:conf/cvpr/DingXLXL19}&	ResNet-101&	20&	86.2\\
	R$^{3}$Det \cite{DBLP:journals/corr/abs-1908-05612}&	ResNet-101&	21&	89.26\\
	DRN	\cite{DBLP:conf/cvpr/PanRSDYGMX20}&	Hourglass104&	-	&92.7*\\
	CenterMap-Net \cite{9151222}&	ResNet-50&	15&	92.8*\\
	\midrule
	\emph{ours:}&&&\\
	TS$^{4}$Net & ResNet-50& 1& 89.44/92.60*\\
	TS$^{4}$Net & ResNet-101& 1& \textbf{89.87}/ \textbf{94.79*}\\
  \bottomrule
\end{tabular}
\end{table}

\textbf{Results on HRSC2016.} HRSC2016 is a challenging dataset, which contains a lot of rotated ships with large aspect ratios. The test results of HRSC2016 are depicted in Table \ref{tab5}. TS$^{4}$Net achieves mAP of 89.44\% (VOC2007) and 92.60\% (VOC2012) with ResNet-50 as backbone. With ResNet-101 backbone, our model achieves 89.87\% and 94.79\% mAP under VOC2007 and VOC2012 metrics, respectively, which outperforms the other methods. The experimental results verify the effectiveness of our method.

\begin{table*}
\centering
\footnotesize
  \caption{Detection accuracy of OBB task on DOTA dataset. R-50 stands for ResNet-50 (likewise for R-101 and R-152), and H-104 represents Hourglass-104.}
  \label{tab4}
\setlength{\tabcolsep}{1.2mm}
  \begin{tabular}{c|c|ccccccccccccccc|c}
    \toprule
    Method	&Backbone &	PL& 	BD& 	BR &	GTF& 	SV &	LV 	&SH 	&TC &	BC 	&ST 	&SBF &	RA &	HA& 	SP &	HC&	mAP\\
    \midrule
	\emph{two-stage:}& &&&&&&&&&&&&&&&&\\
    FR-O \cite{DBLP:conf/cvpr/XiaBDZBLDPZ18}&	R-101&	79.09	&69.12&	17.17&	63.49&	34.20 &	37.16 &	36.20 &	89.19 &	69.60 &	58.96 &	49.40& 	52.52 &	46.69 &	44.80 &	46.30 &	52.93\\
	R-DFPN \cite{DBLP:journals/remotesensing/YangSFYSYG18} &R-101&	80.92 &	65.82 &	33.77& 	58.94 &	55.77 &	50.94 &	54.78 &	90.33 &	66.34& 	68.66 &	48.73& 	51.76& 	55.10 &	51.32 &	35.88 &	57.94\\
	R$^{2}$CNN \cite{DBLP:journals/corr/JiangZWYLWFL17}&R-101	&80.94& 	65.67& 	35.34 &	67.44 &	59.92 &	50.91& 	55.81 &	90.67& 	66.92& 	72.39 &	55.06& 	52.23 &	55.14 &	53.35 &	48.22 &	60.67\\
	RRPN \cite{rrpn} &R-101&	88.52 &	71.20& 	31.66 &	59.30 &	51.85 &	56.19& 	57.25 &	90.81 &	72.84 &	67.38& 	56.69& 	52.84 	&53.08 &	51.94 &	53.58& 	61.01\\
	ICN	\cite{DBLP:conf/accv/AzimiVB0R18}&R-101&	81.36& 	74.30& 	47.70 &	70.32& 	64.89& 	67.82 &	69.98& 	90.76& 	79.06& 	78.20& 	53.64 &	62.90 &	67.02& 	64.17& 	50.23 &	68.16\\
	RoI-Trans. \cite{DBLP:conf/cvpr/DingXLXL19}&	R-101	&88.64 &	78.52 &	43.44 &	\textbf{75.92} 	&68.81& 	73.68& 	83.59 &	90.74 &	77.27& 	81.46 &	58.39 &	53.54 &	62.83 &	58.93 &	47.67 &	69.56\\
	SCRDet	\cite{DBLP:conf/iccv/YangYY0ZGSF19} &R-101	&\textbf{89.98} &	80.65 &	52.09 &	68.36 &	68.36 &	60.32 &	72.41& 	90.85 &	\textbf{87.94} &	\textbf{86.86} &	\textbf{65.02} &	66.68 &	66.25 &	68.24 &	65.21& 	72.61\\
	\midrule
	\emph{one-stage:}& &&&&&&&&&&&&&&&&\\
	R-RetinaNet \cite{DBLP:conf/iccv/LinGGHD17}&	R-50&	89.27	&82.09	&38.09&	70.97	&74.93&	64.04&	78.15&	90.82&	86.40	&76.25&	59.75&	\textbf{67.07}&	54.62	&69.85&	49.49	&70.12\\
	DRN	\cite{DBLP:conf/cvpr/PanRSDYGMX20} &H-104&	88.91 &	80.22& 	43.52& 	63.35& 	73.48& 	70.69 &	84.94 &	90.14 &	83.85& 	84.11 &	50.12 	&58.41& 	67.62 &	68.60 &	52.50 &	70.70\\
	R$^{3}$Det \cite{DBLP:journals/corr/abs-1908-05612}&	R-101&	89.54 &	81.99 &	48.46& 	62.52& 	70.48 &	74.29 	&77.54 	&90.80& 	81.39 &	83.54& 	61.97& 	59.82 &	65.44 &	67.46 &	60.05 &	71.69\\
	BBAVector \cite{bba}&  R-101& 88.35& 79.96& 50.69 &62.18& 78.43& 78.98& \textbf{87.94} &90.85& 83.58& 84.35 &54.13& 60.24 &65.22 &64.28& 55.70& 72.32\\
	R$^{3}$Det \cite{DBLP:journals/corr/abs-1908-05612}&	R-152	&89.49 &	81.17 &	50.53 &	66.10& 	70.92 &	\textbf{78.66} &	78.21 &	90.81& 	85.26 &	84.23& 	61.81 &	63.77 &	\textbf{68.16} &	69.83 &	\textbf{67.17}& 	73.74\\
	\midrule
	\emph{ours:}& &&&&&&&&&&&&&&&&\\
	TS$^{4}$Net	&R-50	&89.33	&83.73	&50.26&	71.33&	\textbf{78.82}&	75.28&	86.31&	90.85&	84.54&	85.51&	63.60&	65.57&	66.51&	72.01&	56.88&	74.70\\
	TS$^{4}$Net	&R-101	&89.59	&\textbf{84.07}	&\textbf{53.41}	&73.23	&78.76&	77.11&	87.04&	\textbf{90.88}	&85.94&	85.48&	63.04&	64.98&	68.05&	\textbf{72.34}&	60.58&	\textbf{75.63}\\
  \bottomrule
\end{tabular}
\end{table*}

\textbf{Results on UCAS-AOD.} The UACS-AOD dataset contains plane and car objects with rotated ground-truth. The experimental results on the UACS-AOD dataset are shown in Table \ref{ucas}. Our model achieves 96.09\% mAP results with ResNet-50, and 96.44\% mAP with ResNet-101 as backbone. Compared to the baseline, the mAP is improved by 2.34\% (96.09\% $v.s.$ 93.75\%). Compared with other methods in Table \ref{ucas}, our model achieves the highest accuracy and the best results under the car category.

\begin{table}
\centering
  \caption{Detection results on  UCAS-AOD dataset.}
  \label{ucas}
  \begin{tabular}{c|c|c|c}
    \toprule
    Method	&mAP&	Plane&	Car\\
    \midrule
	R-DFPN \cite{DBLP:journals/remotesensing/YangSFYSYG18}	&89.20&	95.90&	82.50\\
	DRBox \cite{DBLP:journals/corr/abs-1711-09405}	&89.95&	94.90&	85.00\\
	R-RetinaNet \cite{DBLP:conf/iccv/LinGGHD17}&	93.75&	97.79&	89.71\\
	S$^{2}$ARN \cite{DBLP:journals/access/BaoZZZLL19}&	94.90&	97.60&	92.20\\
	ICN \cite{DBLP:conf/accv/AzimiVB0R18}&	95.67&	-&	-\\
	FADet \cite{DBLP:conf/icip/LiXCWZY19}&	95.71&	\textbf{98.69}&	92.72\\
	R3Det \cite{DBLP:journals/corr/abs-1908-05612}&	96.17&	98.20&	94.14\\
	\midrule
	\emph{ours:}&&&\\
	TS$^{4}$Net-R50 & 96.09	&98.58&	93.60\\
	TS$^{4}$Net-R101 & \textbf{96.44}&	98.42&	\textbf{94.46}\\
  \bottomrule
\end{tabular}
\end{table}

\section{Conclusion}
In this paper, we present a drone-based rotating object detection dataset named UAV-ROD, and believe that our proposed dataset can promote the progress of the rotating object detection community. Besides, we design a rotating object detector TS$^{4}$Net based on the two-stage sample selective strategy. Combined with the anchor refinement module and the two-stage sample selective strategy, the TS$^{4}$Net can address the rotated regression problem of horizontal anchor satisfactorily. Extensive experiments have demonstrated our method can achieve competitive performance against most state-of-the-art methods.

\section{Appendix}

\subsection{Analysis of different two-stage sample seletive strategies.}
\begin{figure}[h]
  \centering
  \includegraphics[width=\linewidth]{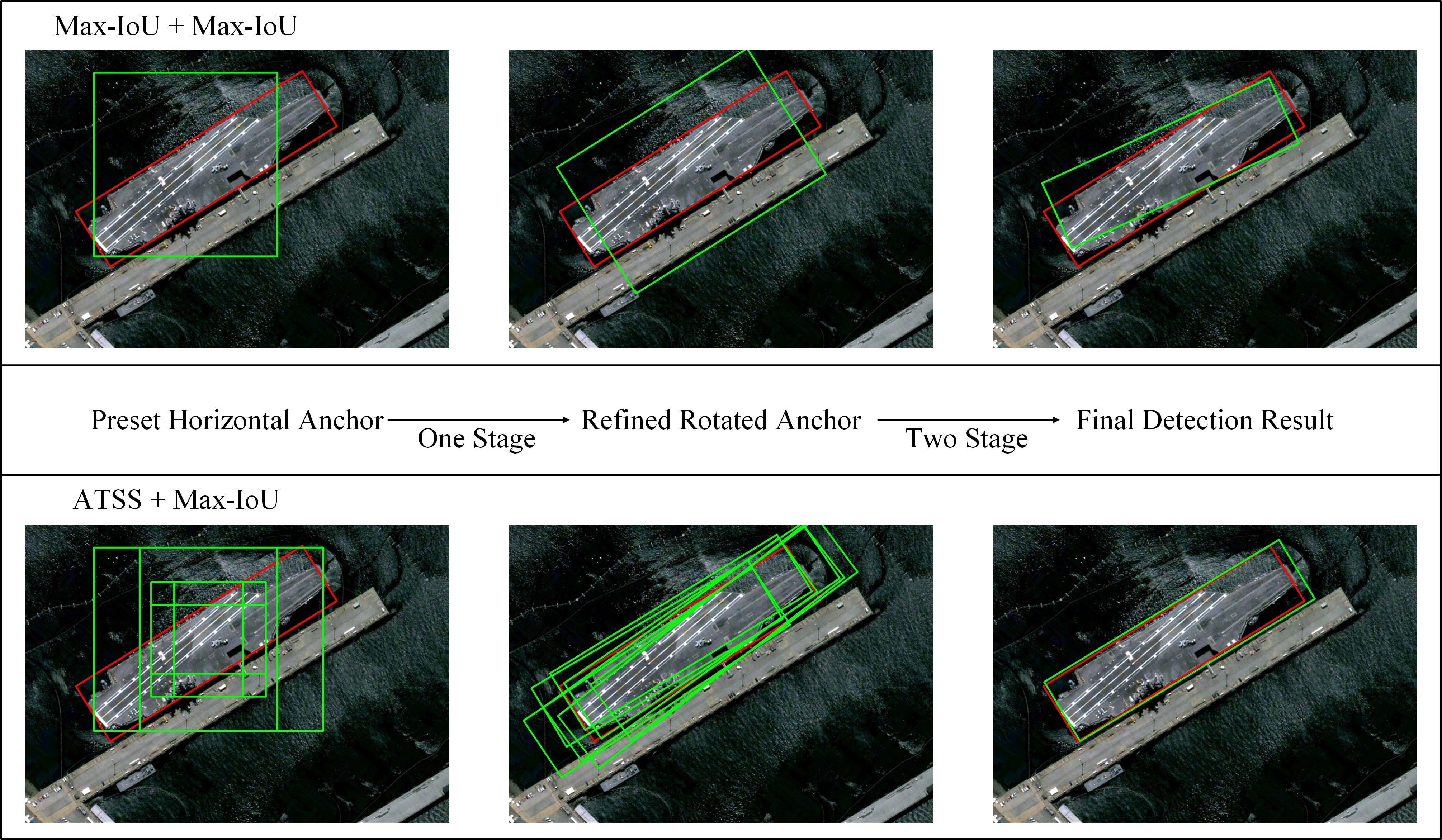}
  \caption{Visualization of different two-stage sample seletive strategy results.}
  \label{tv}
\end{figure}

Figure \ref{tv} shows the positive sample visualization results for the two different strategies. As can be seen from Figure \ref{tv}, the Max-IoU method is more constrained in selecting positive samples, and the number of high-quality positive samples selected is low, making the training inadequate and imbalance. The ATSS+Max-IoU method selects sufficient positive horizontal samples by the center distance information in first-stage, and then selects high-quality rotated anchors for the second regression by the constrained Max-IoU method. This method can select sufficient high-quality samples, thus benefiting the training process.

\subsection{Qualitative Results}
\begin{figure}[h]
  \centering
  \includegraphics[width=\linewidth]{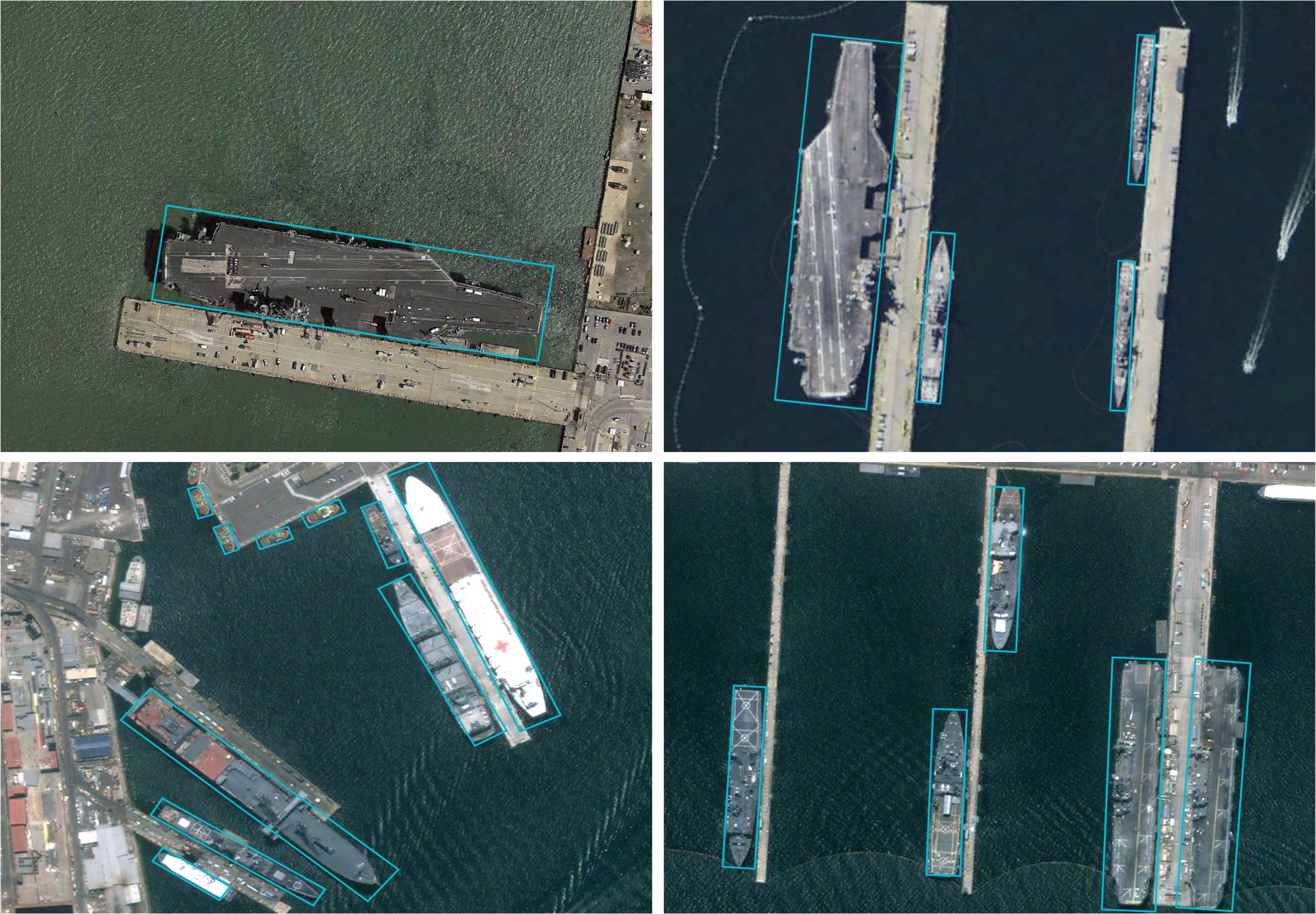}
  \caption{Qualitative results of TS$^{4}$Net on HRSC2016 dataset.}
  \label{hrsc}
\end{figure}

We show some qualitative detection results on different datasets, including UAV-ROD (Figure \ref{uav}), DOTA (Figure \ref{dota}), HRSC2016 (Figure \ref{hrsc}) and UCAS-AOD (Figure \ref{ucas}) datasets.

\begin{figure}[h]
  \centering
  \includegraphics[width=\linewidth]{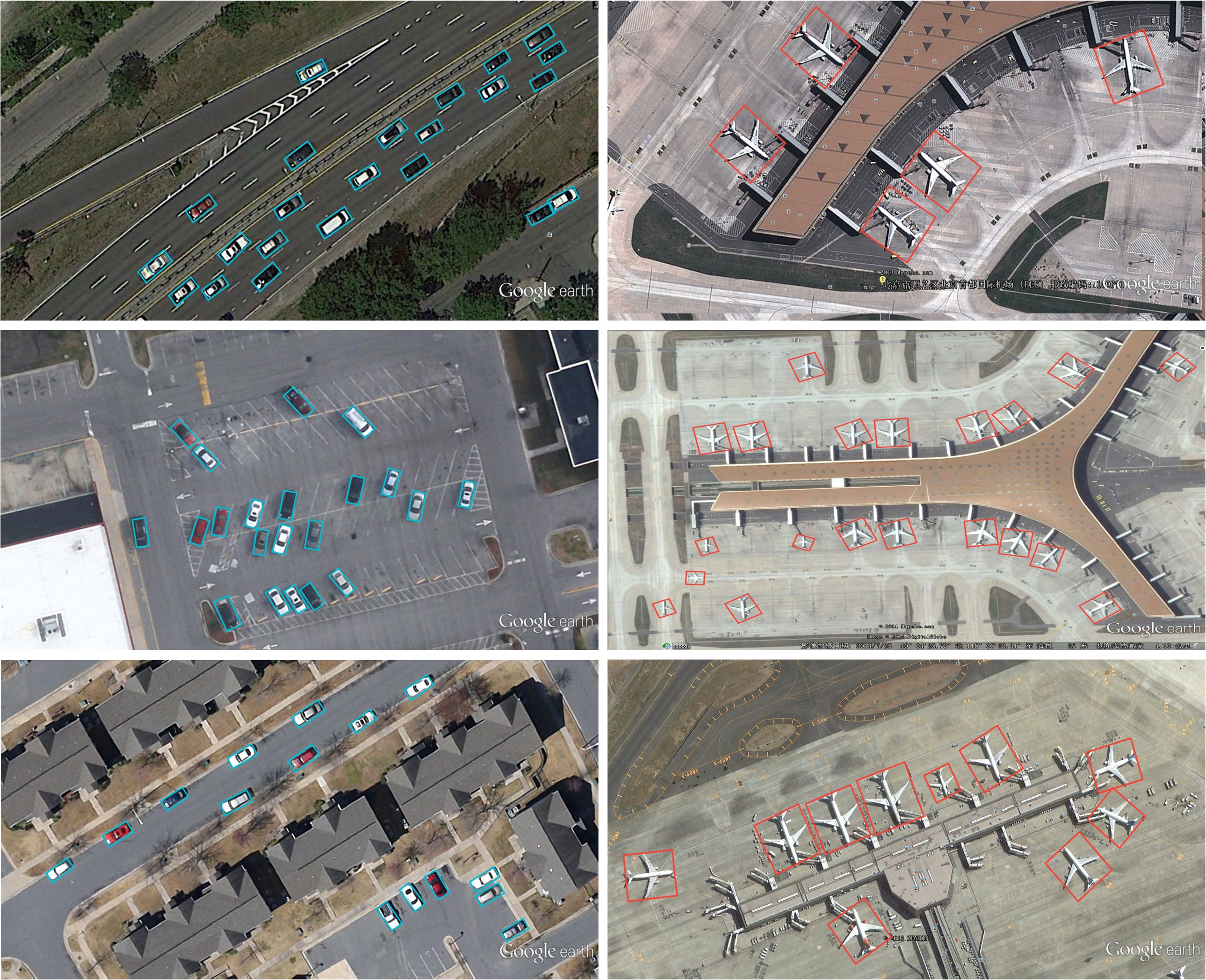}
  \caption{Qualitative results of TS$^{4}$Net on UCAS-AOD dataset.}
  \label{ucas}
\end{figure}

\begin{figure*}
  \centering
  \includegraphics[width=\linewidth]{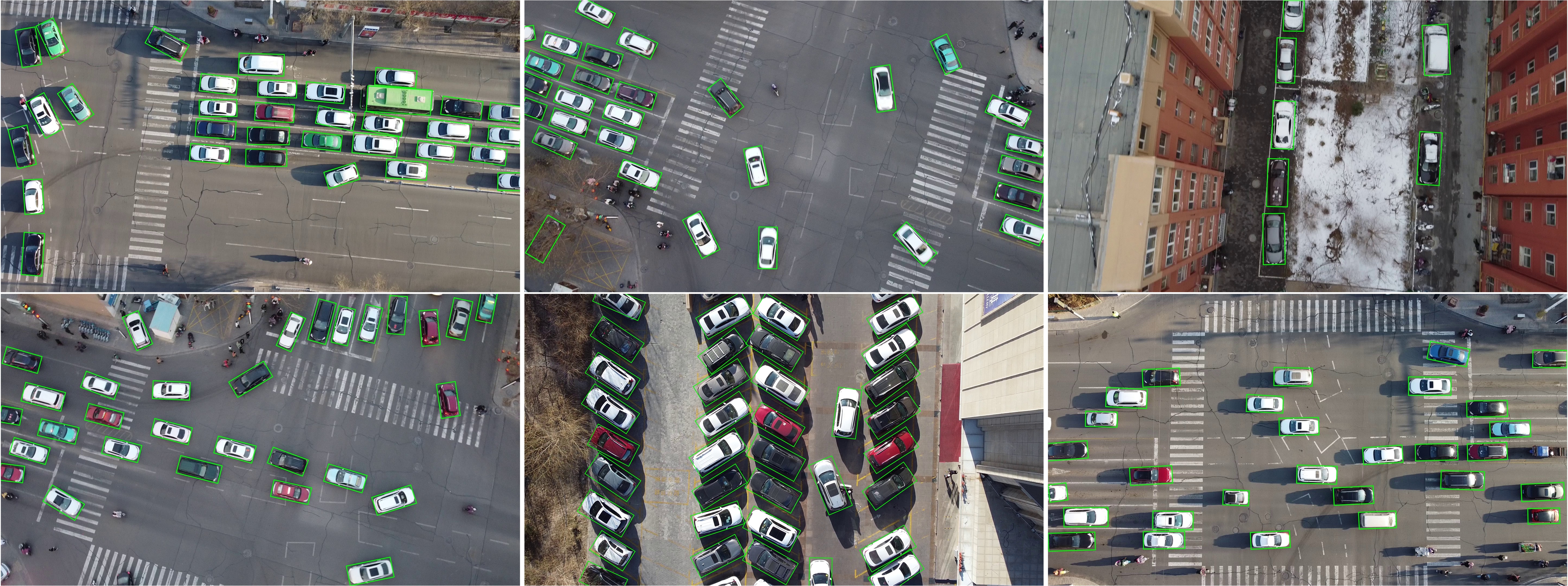}
  \caption{Qualitative results of TS$^{4}$Net on UAV-ROD dataset.}
  \label{uav}
\end{figure*}

\begin{figure*}
  \centering
  \includegraphics[width=\linewidth]{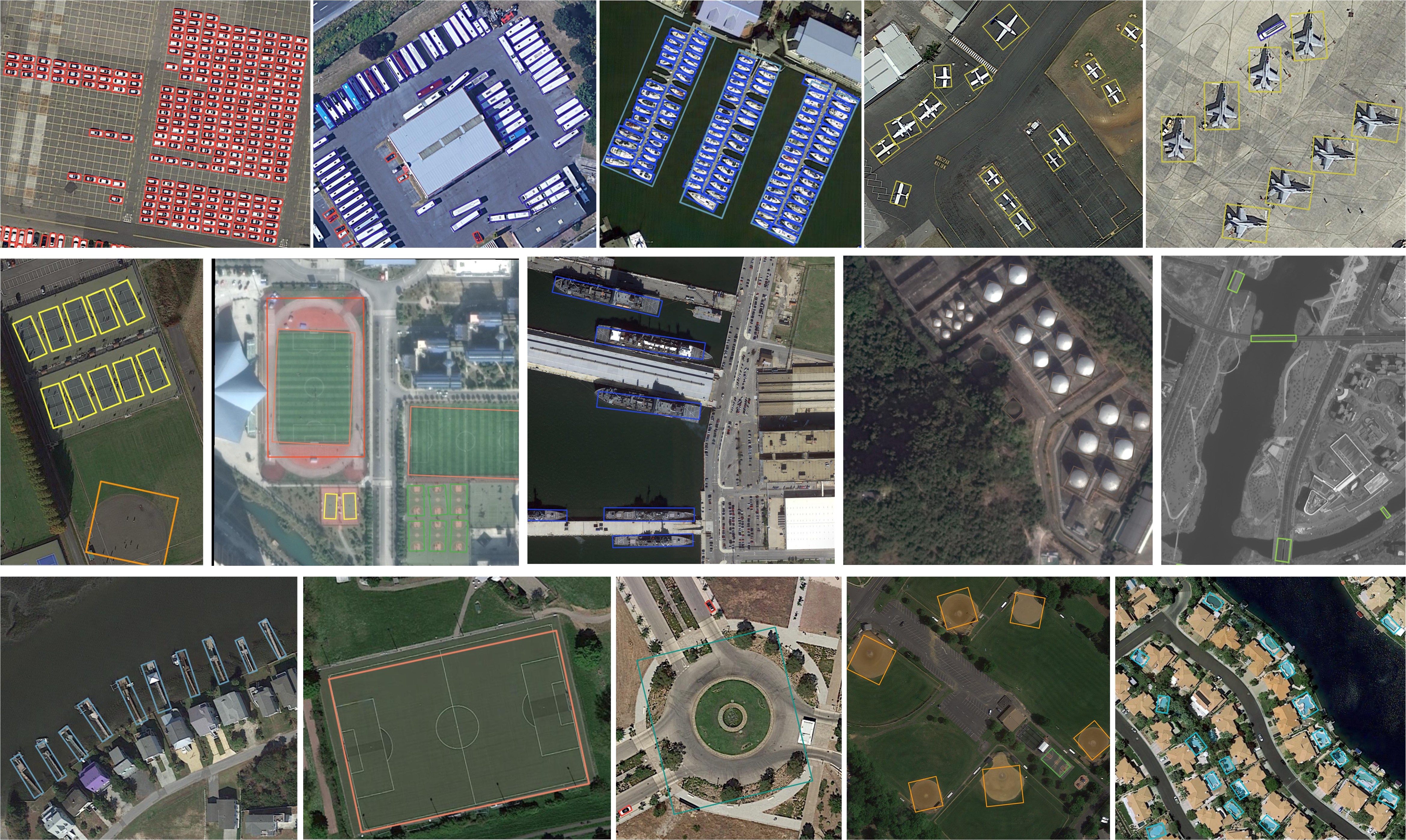}
  \caption{Qualitative results of TS$^{4}$Net on DOTA dataset.}
  \label{dota}
\end{figure*}
%-------------------------------------------------------------------------

{\small
\bibliographystyle{ieee_fullname}
\bibliography{egbib}
}

\end{document}